\documentclass{article}

\usepackage{arxiv}

\usepackage[utf8]{inputenc} 
\usepackage[T1]{fontenc}    
\usepackage{hyperref}       
\usepackage{url}            
\usepackage{booktabs}       
\usepackage{amsfonts}       
\usepackage{nicefrac}       
\usepackage{microtype}      
\usepackage{lipsum}		
\usepackage{graphicx}
\usepackage{natbib}
\usepackage{doi}

\usepackage{alltt}
\usepackage{amsfonts, amsmath, amssymb}
\usepackage[english]{babel}
\usepackage{booktabs}
\usepackage{bm}
\usepackage{eucal}
\usepackage{enumerate}
\usepackage{float}
\usepackage{dsfont}
\usepackage{enumitem}
\usepackage{epsfig}
\usepackage{fontenc}
\usepackage{listings}
\usepackage{lscape}
\usepackage{rotating}
\usepackage[table]{xcolor}
\usepackage{lscape}
\usepackage{mdframed}
\usepackage{natbib}
\usepackage{rotating}
\usepackage{verbatim}
\usepackage{url}
\usepackage[table]{xcolor}

\definecolor{codegreen}{rgb}{0,0.6,0}
\definecolor{codegray}{rgb}{0.5,0.5,0.5}
\definecolor{codepurple}{rgb}{0.58,0,0.82}
\definecolor{backcolour}{rgb}{0.95,0.95,0.92}
\definecolor{grey}{RGB}{128, 128, 128} 

\usepackage{scalerel}
\usepackage{stackengine}
\usepackage{pgf}
\newcounter{iloop}
\newcommand\openbigstar[1][0.7]{%
  \scalerel*{%
    \stackinset{c}{-.125pt}{c}{}{\scalebox{#1}{\color{white}{$\bigstar$}}}{%
      $\bigstar$}%
  }{\bigstar}
}
\newcommand{\Stars}[1]{\ensuremath{%
\pgfmathtruncatemacro{\imax}{ifthenelse(int(#1)==#1,#1-1,#1)}%
\pgfmathsetmacro{\xrest}{0.9*(1-#1+\imax)}%
\setcounter{iloop}{0}%
\loop\stepcounter{iloop}\ifnum\value{iloop}<\the\numexpr\imax+1
\bigstar\repeat
\openbigstar[\xrest]%
\setcounter{iloop}{0}%
\loop\stepcounter{iloop}\ifnum\value{iloop}<\the\numexpr5-\imax\relax
  \openbigstar[.9]\repeat}}

\date{}
\IfFileExists{upquote.sty}{\usepackage{upquote}}{}

\title{Using ChatGPT for Data Science Analyses}

\author{ \href{https://orcid.org/0000-0000-0000-0000}{\includegraphics[scale=0.06]{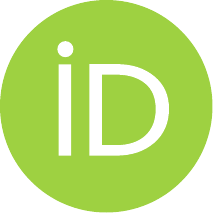}\hspace{1mm}Ozan Evkaya}\thanks{Ozan.Evkaya@ed.ac.uk} \\
	School of Mathematics\\
	University of Edinburgh\\
	\texttt{Ozan.Evkaya@ed.ac.uk} \\
	\And
	\href{https://orcid.org/0000-0000-0000-0000}{\includegraphics[scale=0.06]{orcid.pdf}\hspace{1mm}Miguel de Carvalho} \\
	School of Mathematics \\
	University of Edinburgh\\
	\texttt{miguel.decarvalho@ed.ac.uk} \\
}

\renewcommand{\shorttitle}{\textit{arXiv} Template}

\hypersetup{
pdftitle={Data Analysis in ChatGPT: A Critical Review},
pdfsubject={q-bio.NC, q-bio.QM},
pdfauthor={Ozan Evkaya, Miguel de Carvalho},
pdfkeywords={First keyword, Second keyword, More},
}

\begin{document}
\maketitle

\begin{abstract}
As a result of recent advancements in Generative AI, the field of Data Science is prone to various changes. The way practitioners construct their Data Science workflows
is now irreversibly shaped by recent advancements, particularly by tools like OpenAI’s Data Analysis plugin. While it offers powerful support as a quantitative co-pilot, its limitations demand careful consideration in empirical analyses.

This paper assesses the potential of ChatGPT for Data Science analyses, illustrating its
capabilities for data exploration and visualization, as well as for commonly used supervised
and unsupervised modeling tasks. While we focus here on how the Data Analysis plugin
can serve as a co-pilot for the Data Science workflow, its broader potential for automation
is implicit throughout.
\end{abstract}


\section{Introduction}
\label{intro}
Machines have been key to the production of statistics since at least the 19th century. In a
paper read before the Royal Statistical Society in 1894, Herman Hollerith introduced his invention,
the ‘Tabulating Machine’. Prior to the use of this machine, it took about 10 years to process the
collected data from the 1880 US Census to arrive at a final report. Quoting \cite{hollerith1894}:
\begin{quote}\footnotesize 
``\textsl{While engaged in work in the tenth census, that of 1880, [...] I found, for example, that while we had collected the information regarding the conjugal condition of our 50\,000\,000 inhabitants, we were unable to compile this information even in its simplest form, so that, until the census of 1890, we never even knew the proportion of our population that was single, married, and widowed.}''
\end{quote}

The `Tabulating Machine' was able to process data recorded on punch cards, and thanks to it the 1890 census was completed in only 18 months and within a much smaller budget. Although much has changed since Hollerith’s invention, there was no turning back: The use of machines for handling and modeling data continued throughout the 20th century and has been increasing at a
steep rate in the 21st century with the rise of AI.

This note introduces the latest contribution to that space: OpenAI’s Data Analysis (DA) extension for ChatGPT. We provide an outlook of its performance across various tasks; the walkthrough covers initiating the software and assessing its capabilities in exploring, visualizing, and extracting insights from data. ChatGPT’s DA provides a comprehensive Python-based toolbox, supports
memory-intensive operations, and leverages the strengths of LLMs (Large Language Models) \cite{Mollick2023}.  Also, it incorporates a Python interpreter into the ChatGPT environment for on-the-fly
code execution, and to ensure maximum security all computing occurs within a firewall-protected
sandbox.

The growing interest in automating Data Science is well documented (e.g., \cite{Bie2022},
and references therein), and tools like the DA plugin offer strong potential to streamline workflows.
Below, we illustrate how it can act as a co-pilot for Data Science tasks. Although this is not our
main focus, its broader potential for automation is implicit and worth keeping in mind. 

Although LLMs remain prone to hallucinations, biases, and inaccuracies, they nonetheless have
the potential to become central tools in the statistical software landscape; indeed, the significant
promise of LLMs in Data Science has been highlighted recently by \cite{Xinming2023}. In particular,
DA enables even non-programmers to run code and obtain elaborate analyses. Despite significant
progress, we argue that it is key to emphasize that no AI-powered statistical software should operate
without human critique and oversight, nor should it be considered a complete substitute for the
skills of a professional data analyst.


Both supervised and unsupervised learning tasks will be covered in this review; for technical
details on the methods covered here see, for instance, \cite{murphy2022, murphy2023}. A publicly available \texttt{laptop} price dataset from Kaggle \cite{Kushwaha23} is used for exploratory data analysis
and visualizations, while the publicly available \texttt{duke\_forest} dataset from the \textit{openintro} R package \cite{openintro23} is employed for the modeling examples. While leading the conversation with DA, we primarily focused on simple prompts and on following the suggestions provided
by DA.


We start with exploratory analyses and visualizations in Section 2. The next sections are dedicated to supervised and unsupervised learning tasks. Lastly, the paper ends with some closing
remarks. All statistical analyses and comments below in bluish white boxes, figures, and code were
generated by ChatGPT’s DA. Throughout, we will use a star rating system to score statements
and analyses produced by DA. In addition to this rating system, we briefly highlight the impact of
prompting and repeated prompting over some tasks.

\section{Seeing Through Data}\label{exploratory}
\subsection{Getting Started}
\begin{figure}[H]
\centering
\includegraphics[scale=0.4]{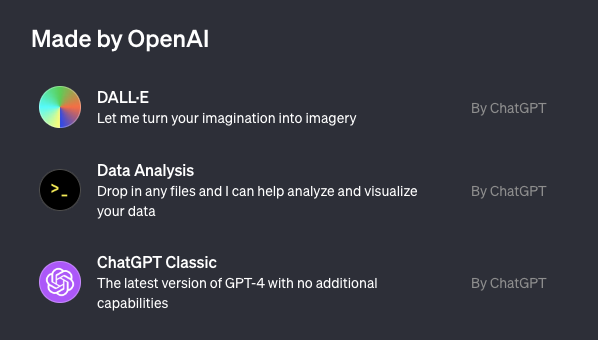}
\caption{\footnotesize ChatGPT window to turn on the Data Analysis feature.}
\label{fig:GPT_DA}
\end{figure}

To assess the strengths and weaknesses of ChatGPT's DA capabilities we start with the \texttt{laptop} dataset. This dataset is provided in the widely-used CSV format, enabling easy upload and prompt interaction with the DA console. Fig~\ref{fig:GPT_DA} shows how the DA plugin can be activated within the ChatGPT interface.

After uploading the data, the user can immediately ask data analysis-related questions to the DA extension. In a seamless way, DA translates such queries into Python code and provides answers in text---after running the code under the hood.

\subsection{Loading and Preprocessing Data}
Loading a dataset in DA is as easy as dragging and dropping a CSV file into the console. Subsequently, DA assists in outlining the steps necessary for the planned analysis. The prompt ``We have attached a dataset and we need to do some data analysis project with it. Can you create a report on this dataset?''
immediately led to an excellent list of tasks to follow: 

\begin{mdframed}[backgroundcolor=grey!20, linewidth = 0pt] \vspace{0.1cm} \footnotesize 
  \textbf{ChatGPT's DA} \strut \hfill \Stars{5}  \\ 
  {Great, we'll conduct the following analyses on your dataset:}

\begin{enumerate}[itemsep=0pt, parsep=0pt, topsep=0pt, partopsep=0pt]
    \item {Descriptive statistics for numerical features.}
    \item {Distribution of laptops by company, type, and other categorical features.}
    \item {Price distribution and its relationship with other features.}
    \item {Analysis of the impact of TouchScreen, IPS display, and PPI on the price.}
    \item {The distribution of laptop features by operating system.}
\end{enumerate}

\noindent {We will start with the descriptive statistics for the numerical features. Let's proceed with this analysis.} 
{Here are the descriptive statistics for the numerical features in the dataset:} \vspace{-.4cm}
\begin{center}
  $[$...$]$
\end{center}
\end{mdframed}

\begin{figure}[H]
\centering
\includegraphics[scale=0.4]{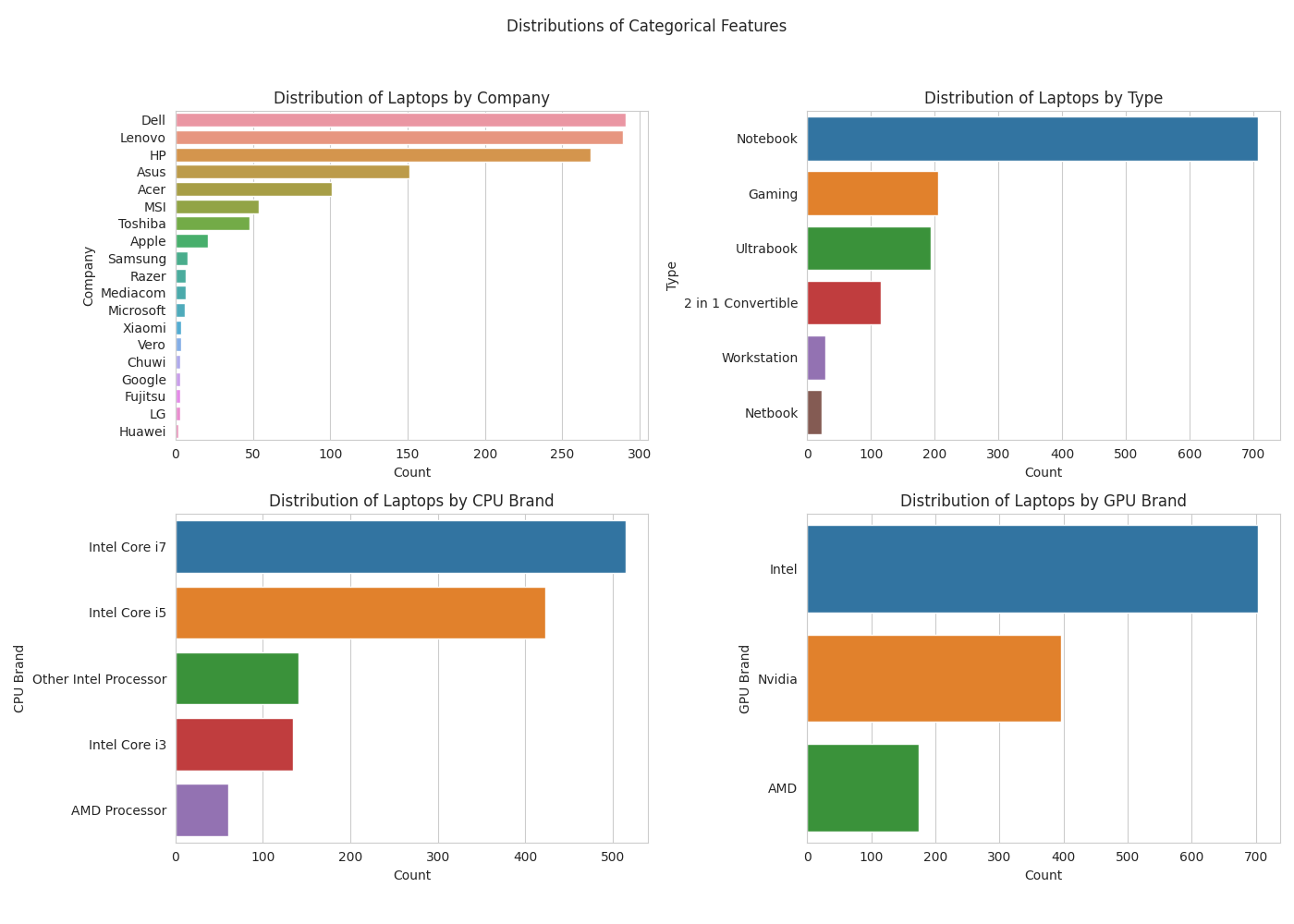}
\caption{\footnotesize \texttt{laptop} data: Frequency per company, type, and CPU--GPU brands.} \footnotesize 
\Stars{5}
\label{fig:Dist_Categs}
\end{figure}

\noindent After producing a battery of summary statistics, DA suggested other steps such as:

\begin{mdframed}[backgroundcolor=grey!20, linewidth = 0pt] \vspace{0.1cm} 
\footnotesize 
  \textbf{ChatGPT's DA} $[$...$]$ \strut \hfill \Stars{5}  \\ {Next, we can look at the distribution of laptops by company, type, and other categorical features. Shall we proceed?}
\end{mdframed}
\noindent
after which we answered ``Yes.'' This resulted in the automatic generation of Fig~\ref{fig:Dist_Categs}, which sets the stage for the subsequent discussion on visualizations in the next section.

In addition to the above CSV file, we tested various datasets using the GPT DA interface and
API. Uploading XLSX and TXT files is straightforward, but issues arise with formats such as
RDATA, which require extra conversion (e.g., via \texttt{pyreadr}). Similar problems occur with SAV
and SASHDAT files, as SPSS and SAS formats are not yet supported, making manual conversion
necessary. Dataset size also poses challenges, as large files can slow down importing and parsing.
Overall, based on our experiences, CSV, XLSX, and TXT remain the most practical formats.

\subsection{Data Visualization} 
One of the powerful aspects of the DA extension is that it excels at generating data visualizations and seamlessly executing successive analysis steps.

In our first visualization, shown in Fig~\ref{fig:Dist_Categs}, DA created a set of bar plots summarizing the frequency of laptops by company, type, and CPU--GPU brands. As a next move, DA suggested: 

\begin{mdframed}[backgroundcolor=grey!20, linewidth = 0pt]
\vspace{0.1cm} \footnotesize 
\textbf{ChatGPT's DA} \strut \hfill \Stars{5}  \\  $[$...$]$ look at the price distribution and its relationship with other features. Shall we continue with that analysis? 
\end{mdframed}

\begin{figure}[H]
\centering
\includegraphics[scale=0.4]{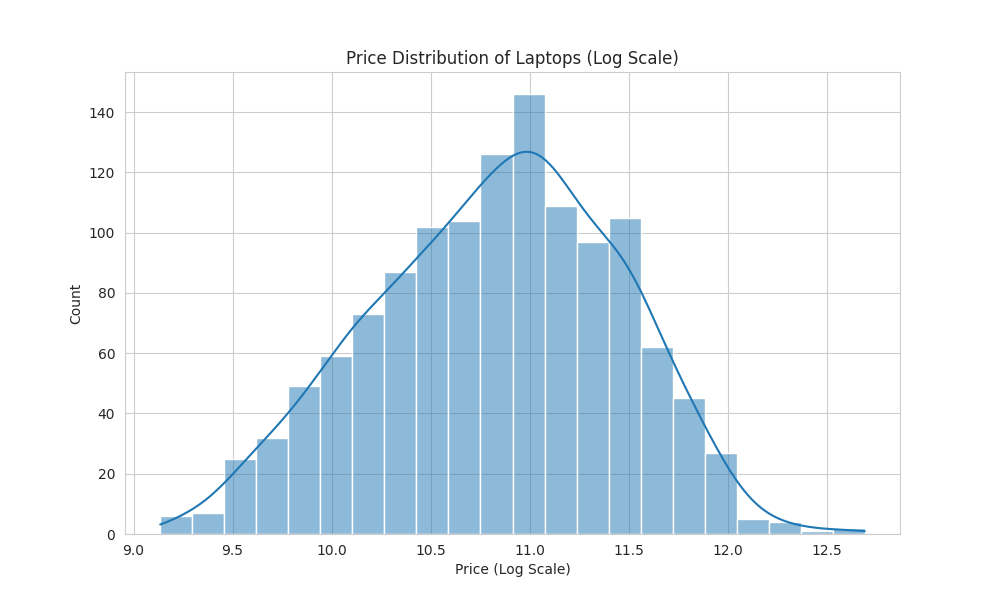}
\caption{\footnotesize Histogram of \texttt{laptop} prices.}
\footnotesize
\Stars{2}
\label{fig:priceDist_hist}
\end{figure}

\noindent After we answered “yes,” DA produced Fig~\ref{fig:priceDist_hist}; yet, it incorrectly suggested that prices are depicted
on a log scale (cf x-axis), whereas closer inspection of the raw CSV file revealed that this is misleading as prices were actually depicted on the original scale. Next, the Data Analysis plugin also
suggested and generated several additional visualizations, including boxplots and bar charts. For
instance, it visualized prices across categories using boxplots, but alternatively used bar charts to
display average prices per category, which is not an appropriate representation of the underlying
distribution

\begin{figure}[H]
\centering
\includegraphics[scale=0.4]{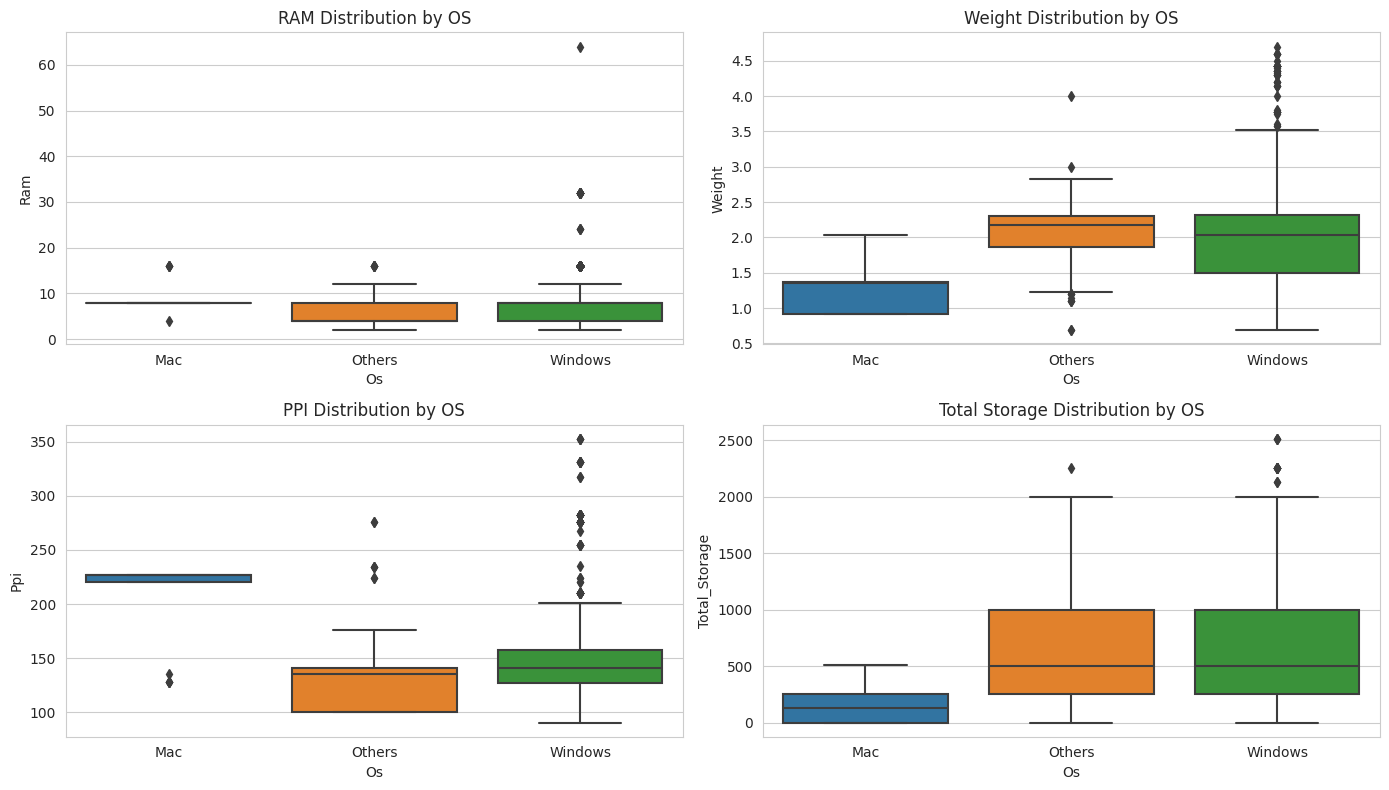}
\caption{\footnotesize Side-by-side boxplots for \texttt{laptop} data.}
\footnotesize
\Stars{5}
\label{fig:priceDist_boxplot}
\end{figure}

\noindent To wrap up: With the exception of Fig~\ref{fig:priceDist_hist}, the DA plugin performed satisfactorily in exploring
and visualizing data. However, given the random nature of DA, repeated use of the same prompt
may still yield inconsistently labeled histograms. As a further challenge, we next turn to another
specific data visualization task.
\subsection{{Mosaic Plot}}
\begin{figure}[H]
\centering
\includegraphics[scale=0.375]{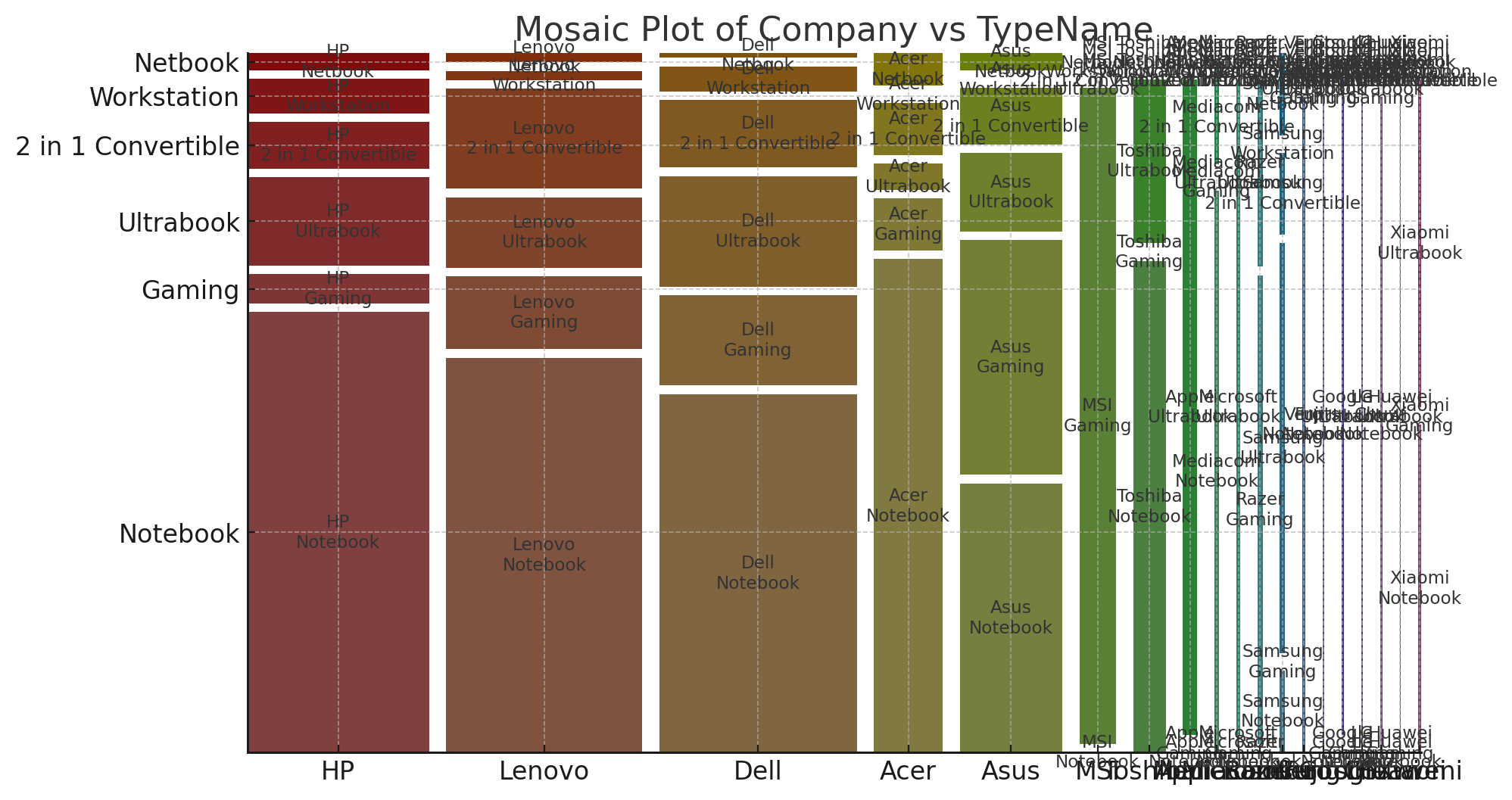}
\caption{\footnotesize Mosaic plot of \texttt{Company} vs \texttt{TypeName}.}
\footnotesize
\Stars{3}
\label{fig:mosaic_plot}
\end{figure}

{As another example, we considered sketching mosaic plots. In response to repeated prompts, the DA plugin primarily relied on the same set of categorical variables (e.g., \textit{Company} or \textit{TypeName}), but often produced visualizations with limited readability due to suboptimal color choices or poorly positioned labels. One such example can be seen in Figure  \ref{fig:mosaic_plot}.}

The tested prompt was: ``For the attached dataset, can you create a mosaic plot for the categorical variables?'' Clearly, this was a bad prompt and ChatGPT's first response was good given the circumstances; visual details were, however, not quite satisfactory, and depending on the various categorical variables, text labels were overlapping, and repeated labels avoided readability. But again on the positive side, DA responded with a follow-up question to the user for possible resolutions, such as:\vspace{0.2cm}

\begin{mdframed}[backgroundcolor=grey!20, linewidth = 0pt]
\vspace{0.1cm} \footnotesize 
\textbf{ChatGPT's DA} \strut \hfill \Stars{5}  \\  $[$...$]$ {Would you like me to:}

{- Focus on a subset of two variables (e.g., Company and TypeName)?} 

{- Handle missing combinations and retry the three-variable mosaic?’} 
\end{mdframed}

\noindent {To conclude, most of the generated mosaic plots required some fine-tuning, reinforcing the need of a human-in-the-loop approach.}
\subsection{{Correlation Plot}}

{We also explored the correlation heatmap on the same data set, as a part of our general unclear prompt ``Do all your suggestions'' after getting the suggestion below:} \vspace{0.2cm}

\begin{mdframed}[backgroundcolor=grey!20, linewidth = 0pt] \vspace{0.1cm} 
\footnotesize 
  \textbf{ChatGPT's DA} $[$...$]$ \strut \hfill \Stars{5}  \\ {Next, we can explore relationships between Price and other features or analyze categorical variables like Company or TypeName. Which direction would you like to explore further?}
\end{mdframed}\vspace{0.2cm} \vspace{0.2cm}
\noindent

{Among the created data visualization outputs to represent the anticipated relationship between \textit{Price} vs. other variables, DA displayed a correlation heatmap to understand feature interdependencies. Although the created correlation heatmap seems reasonable from a data vizualization perspective, the interpretation of the results often did not match the generated outcomes. The numerical values in the heatmap were not consistently reflected in DA’s subsequent interpretations.} 

\begin{figure}[H]
\centering
\includegraphics[scale=0.375]{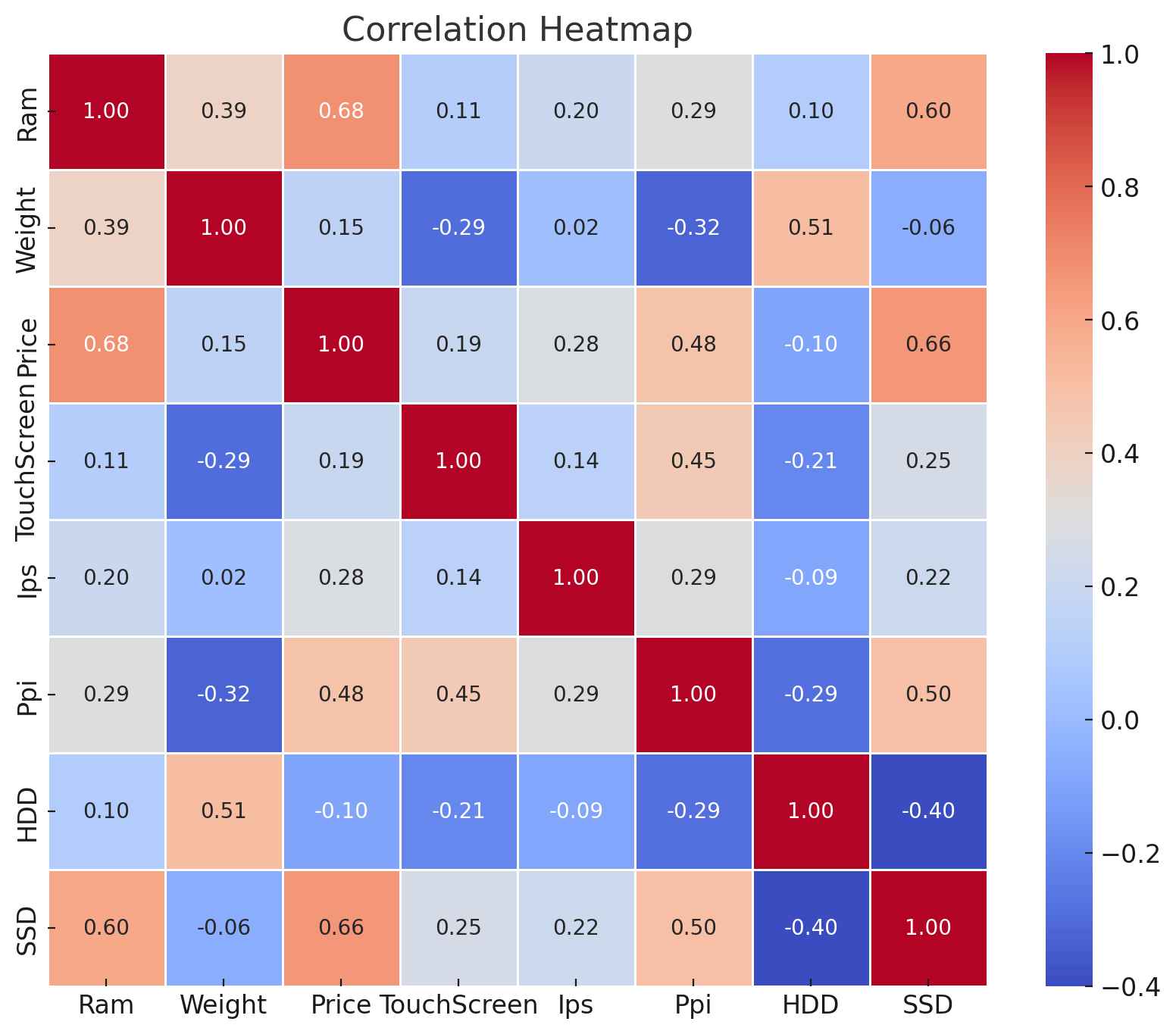}
\caption{\footnotesize Correlation heatmap plot of \texttt{Price} vs. other selected variables.}
\footnotesize
\Stars{4}
\label{fig:correlation_plot}
\end{figure}

{As given in Figure~\ref{fig:correlation_plot}, the output plot with numerical correlation values seems acceptable. One issue is that the nature of some variables is not directly suitable for the correlation analysis such as repeated categorical numbers in HDD or SSD columns. The most important problem is the mismatch between the presented values in Figure~\ref{fig:correlation_plot} and interpreted comments by DA: } \vspace{0.2cm}

\begin{mdframed}[backgroundcolor=grey!20, linewidth = 0pt] \vspace{0.1cm} 
\footnotesize 
\textbf{ChatGPT's DA} $[$...$]$ \strut \hfill \Stars{1}  \\ {SSD (0.64) and RAM (0.74) have strong positive correlations with Price, indicating that storage and memory are key drivers of price. Weight (0.21) and PPI (0.38) show moderate positive correlations with Price.}
\end{mdframed}\vspace{0.2cm} \vspace{0.2cm}
\noindent

\noindent {For example, the correlation between \texttt{Price} and \texttt{Ram} in Figure~\ref{fig:correlation_plot} is 0.68, whereas it is reported as 0.74 in the response above.}

\section{Experiments on Supervised Data}\label{supervised}

\subsection{Warm-up}\label{warmup}
\subsubsection*{First Insights}

In line with our strategy from the previous section, we aimed to keep the complexity
of the prompts as simple as possible—by breaking down the analysis into smaller steps—and by
following the suggestions provided by the DA response. We now consider the \texttt{duke\_forest} dataset
from the \textit{openintro} R package. We kicked off with the question:
 ``We have attached a dataset and we need to do some data analysis project with it. Can you create a report on this data analysis for us?'' Interestingly, the response below indicates that DA is aware of most variables' units---even though these were not detailed in the uploaded CSV file: 
 \begin{mdframed}[backgroundcolor=grey!20, linewidth = 0pt] \vspace{0.1cm}
   \footnotesize 
\textbf{ChatGPT's DA} $[$...$]$ \strut \hfill \Stars{5}  \vspace{-0.4cm}\\ 
\begin{itemize}[itemsep=0pt, parsep=0pt, topsep=0pt, partopsep=0pt]
\item {Unnamed: A seemingly redundant or index column.}
\item {Address: Physical address of the property.}
\item {Price: Sale price of the property.}
\item {Bed: Number of bedrooms.}
\item {Bath: Number of bathrooms.}
\item {Area: The area of the property in square feet.}
\item {Type: Type of property (e.g., Single Family).}
\item {Year Built: The year in which the property was built.}
\item {Heating: Type of heating system in the property.}
\item {Cooling: Type of cooling system.}
\item {Parking: Parking facilities available.}
\item {Lot: Size of the lot in acres.}
\item {HOA: Homeowners Association fees (if any).}
\item {URL: URL for the property listing.}
\end{itemize}
\end{mdframed}

\noindent DA then continued the reply with a few suggestions for a data analysis:  

\begin{mdframed}[backgroundcolor=grey!20, linewidth = 0pt] \vspace{0.1cm}
  \footnotesize 
\textbf{ChatGPT's DA} $[$...$]$ \strut \hfill \Stars{4} \vspace{-0.4cm} \\ 
\begin{itemize}[itemsep=0pt, parsep=0pt, topsep=0pt, partopsep=0pt]
\item {Data Cleaning: Handle missing values, remove unnecessary columns, and correct data types if needed.}
\item {Descriptive Statistics: Summarize the main features of the data.}
\item {Price Analysis: Explore the distribution of property prices and how they relate to other features like area, bedrooms, etc.}
\item {Property Type Analysis: Analyze different property types and their characteristics.}
\item {Temporal Analysis: Look at the year of construction and its impact on price, if the data allows.}
\item {Geographical Analysis: If the data on locations (address) is detailed enough, we can explore geographical trends.}
\item {Correlation Analysis: Determine the relationships between different numerical features.}
\end{itemize}
\end{mdframed}

\noindent Some of these suggestions, such as Geographical Analysis, are not directly applicable due to data limitations; yet, DA acknowledges this possibility. Users can follow the suggested actions and create simple prompts---such as ``Can you give us the descriptive statistics?''---after which DA successfully generates a variety of relevant summaries, including the frequency, mean, standard deviation, and min-max for the numerical variables. As a follow-up, we asked: ``Can we get the correlation analysis for the dataset?'' After this DA computed, among other things, the correlation between Price and Area, which is 0.67, suggesting a strong positive relationship (i.e., as the area of a property increases, its price tends to increase). It also offered some comments on the interpretation of correlation:

\begin{mdframed}[backgroundcolor=grey!20, linewidth = 0pt] \vspace{0.1cm}
  \footnotesize
  \textbf{ChatGPT's DA} \strut \hfill \Stars{3}  \\ {The correlation analysis provides insights into how different numerical variables in your dataset are related to each other. The values range from -1 to 1, where:}
  \begin{itemize}[itemsep=0pt, parsep=0pt, topsep=0pt, partopsep=0pt]
  \item 1 indicates a perfect positive correlation (as one variable increases, the other increases).
  \item -1 indicates a perfect negative correlation (as one variable increases, the other decreases).
  \item 0 indicates no correlation.\vspace{0.1cm}
  \end{itemize} 
\end{mdframed}

\noindent DA is however silent regarding which correlation metric was actually computed. Indeed, the statistical literature encompasses a variety of correlation coefficients, such as Kendall's, Pearson's, and Spearman's correlations, with researchers continually developing new methods \cite{chatterjee2021}. After careful examination of the function executed in the background, we were able to confirm that it computed Pearson's correlation coefficient. Also, the reply by DA should perhaps have emphasized that: \textit{i}) Pearson correlation only detects linear associations, not nonlinear ones; \textit{ii}) correlation does not imply causality.

\subsubsection*{Modeling}
Moving forward, we asked ``We need to do a price analysis.'' Following this prompt, DA generated various suggestions, without ever referring to modeling: 

\begin{mdframed}[backgroundcolor=grey!20, linewidth = 0pt] \vspace{0.1cm}
  \footnotesize 
\textbf{ChatGPT's DA} $[$...$]$ \strut \hfill \Stars{4}  \vspace{-0.4cm} \\ 
  \begin{itemize}[itemsep=0pt, parsep=0pt, topsep=0pt, partopsep=0pt]
\item {Price Distribution: Analyzing the overall distribution of prices to understand the range and common price points.}
\item {Price Trends by Property Features: Investigating how price varies with different features like the number of bedrooms (bed), bathrooms (bath), area, lot size, and year built.}
\item {Price Comparison by Heating and Cooling Types: Since different heating and cooling systems might affect property values.}
\item {Geographical Price Trends: If the address data allows, we can explore how prices vary across different locations.}
\item {Price Trends Over Time: If there's a timestamp associated with each sale (not apparent in the dataset), we could analyze how prices have changed over time.}
\end{itemize}
\end{mdframed}

\noindent To keep the conversation flowing, and for the sake of simplicity, we have opted for Price Distribution. The generated output produced some further sensible exploratory outputs (not shown) and indicated a potential next step:

\begin{mdframed}[backgroundcolor=grey!20, linewidth = 0pt] \vspace{0.1cm}
  \footnotesize
\textbf{ChatGPT's DA} \strut \hfill \Stars{3}  \\ 
{Next, let's look at the relationship between property price and other features such as the number of bathrooms, the area, and the lot size. Would you like to proceed with these analyses?}
\end{mdframed}

\noindent We answered ``Yes, please proceed.'' Although the output included interesting scatterplots---like price vs.~area and price vs.~lot size---it was disappointing that no algorithmic or statistical modeling was conducted at this stage. Hence, next we asked directly ``Which type of modeling is suitable for this dataset?'' This resulted in a long list of sensible models along with an opening remark:

\begin{mdframed}[backgroundcolor=grey!20, linewidth = 0pt] \vspace{0.1cm}
  \footnotesize 
  \textbf{ChatGPT's DA} \strut \hfill \Stars{3}  \\ 
{For this dataset, starting with a multiple linear regression model would be a good approach to understand the linear relationships between price and other variables. If the relationships are nonlinear or if there are significant interactions between variables, more complex models like random forests or gradient boosting machines might be more suitable.}

{Before proceeding with any modeling, it's important to ensure that the data is preprocessed correctly (handling missing values, encoding categorical variables, etc.) and to split the dataset into training and testing sets to evaluate the model's performance.}
\end{mdframed}

\noindent On the positive side, the initial portion of DA’s response is sufficiently detailed for a novice to
grasp the subsequent steps. Yet, it failed to critique the linear regression model, which could yield
negative price predictions. Although this oversight may stem from a similar incorrect assumption
to that made for the laptop data set—that prices were on the log scale—and perhaps DA was
implicitly relying on the same assumption here, it nonetheless remains a notable shortcoming in
the analysis.

The remainder portion of DA's response includes suggestions that look reasonable at first glance, including regression analysis and its potential competitors---like neural network models:  

\begin{mdframed}[backgroundcolor=grey!20, linewidth = 0pt] \vspace{0.1cm}
  \footnotesize 
\textbf{ChatGPT's DA} $[$...$]$ \strut \hfill \Stars{4}  \vspace{-0.4cm} \\
\begin{itemize}[itemsep=0pt, parsep=0pt, topsep=0pt, partopsep=0pt]
    \item {Linear Regression Models}\strut \hfill  $[$...$]$ 
    \item {Decision Tree Regression} \strut \hfill $[$...$]$ 
    \item {Random Forest Regression}\strut \hfill $[$...$]$ 
    \item {Gradient Boosting Machines} \strut \hfill $[$...$]$ 
    \item {Support Vector Machines (SVM) for Regression}\strut \hfill  $[$...$]$ 
    \item {Neural Networks} \strut \hfill $[$...$]$  
\end{itemize}
\end{mdframed}

\noindent Not all users may be familiar with these models, and it can be problematic if they are used without a proper understanding. Thus, it would be beneficial if DA could ask whether the user requires a reference, such as a textbook, to review the details of these methods or if they are already familiarized with them. 

\noindent In terms of detecting the proper model choice, it is highly likely that DA can suggest reasonable
starting points for the novice learners, whereas the implementation of the suggested model or
interpretation of the outcomes are still prone to be misleading.

\subsection{Regression, I: From Linear to Nonlinear Models}
\subsubsection*{Linear Regression}
For the rest of the conservation, some of the models earlier suggested by DA were implemented by prompting, ``Can you implement a linear regression based on your suggestion?'' The response offered a satisfactory roadmap of the analysis:

\begin{mdframed}[backgroundcolor=grey!20, linewidth = 0pt] \vspace{0.1cm}
  \footnotesize 
\textbf{ChatGPT's DA} $[$...$]$ \strut \hfill \Stars{4}  \vspace{-0.4cm} \\ 
\begin{itemize}[itemsep=0pt, parsep=0pt, topsep=0pt, partopsep=0pt]
\item {Preprocess the Data: Handle missing values, if any. Convert categorical variables into numerical formats, if necessary. Select relevant features for the regression model.}
\item {Split the Data: Divide the data into training and testing sets.}
\item {Build the Linear Regression Model: Train the model on the training set. Evaluate the model on the testing set.}
\item {Model Evaluation: Assess the model's performance using metrics such as R-squared, Mean Absolute Error (MAE), Mean Squared Error (MSE), or Root Mean Squared Error (RMSE).}
\end{itemize}
\end{mdframed}

\noindent However, the reply does not mention mainstream model diagnostics such as the residual plots.
Although these are discussed later, their relevance to linear regression models would warrant earlier
mention. 

Overall, the initial steps provided reasonable guidance for a novice learner. The created model
used various predictors to construct a multiple linear regression model:

\begin{mdframed}[backgroundcolor=grey!20, linewidth = 0pt] \vspace{0.1cm}
  \footnotesize 
\textbf{ChatGPT's DA} $[$...$]$\strut \hfill \Stars{4}  \\ {We'll focus on numerical variables like `bed', `bath', `area', `year\_built', and `lot' as predictors for `price'}.
\end{mdframed}

\noindent While the computations were interrupted due to an unexpected error in the platform, the provided code (not shown) implemented the methods accurately, and it could be directly pasted into Jupyter Notebooks or Google Colab \cite{bisong2019}. 
Also, the resulting suggestions were informative and may be instructive for a novice learner:  

\begin{mdframed}[backgroundcolor=grey!20, linewidth = 0pt] \vspace{0.1cm}
  \footnotesize 
\textbf{ChatGPT's DA} $[$...$]$\strut \hfill \Stars{5} \\
To improve the model, you might consider:
\begin{itemize}[itemsep=0pt, parsep=0pt, topsep=0pt, partopsep=0pt]
\item Adding more relevant features, if available. 
\item Trying more complex models that can capture nonlinear relationships.
\item Conducting feature engineering to derive new informative features.
\end{itemize}
\end{mdframed}

\subsubsection*{Other Regression Models}
\noindent Next, we asked ``Can you to fit more complex models?'' DA suggested the following models along with some brief justifications:

\begin{mdframed}[backgroundcolor=grey!20, linewidth = 0pt] \vspace{0.1cm}
  \footnotesize 
\textbf{ChatGPT's DA} \strut \hfill \Stars{4} \vspace{-0.4cm} \\ 
\begin{itemize}[itemsep=0pt, parsep=0pt, topsep=0pt, partopsep=0pt]
\item Decision Tree Regression: Can model complex, non-linear relationships.
\item Random Forest Regression: An ensemble method that uses multiple decision trees.
\item Gradient Boosting Regression: Builds trees one at a time, where each new tree helps to correct errors made by the previous one.
\item Support Vector Regression (SVR): Can be used for both linear and non-linear data.
\end{itemize}
\end{mdframed}

\noindent The previous response was completed as follows:

\begin{mdframed}[backgroundcolor=grey!20, linewidth = 0pt] \vspace{0.1cm}
  \footnotesize 
  \textbf{ChatGPT's DA} $[$...$]$\strut \hfill \Stars{3} \vspace{-0.4cm} \\
\begin{itemize}[itemsep=0pt, parsep=0pt, topsep=0pt, partopsep=0pt]
\item The SVR performed poorly with this dataset, as indicated by the negative $R^{2}$ value, which means it did worse than a simple average model.
\item Given these results, the Random Forest Regressor seems to be the most suitable model for this dataset among the ones tested. It balances complexity and performance, handling the nonlinearities and interactions in the data effectively.
\item Further improvements could be achieved by tuning the model's hyperparameters, feature engineering, or including additional relevant variables if available.
\end{itemize}
\end{mdframed}

At first sight, it may seem surprising that SVR leads to a negative $R^{2}$ value, but it is well known that this can happen for nonlinear models. Quoting \cite{cameron1997}:

\begin{quote}\footnotesize 
  ``\textsl{Application of this measure [$R^2$] to nonlinear models generally
  leads to a measure that can lie outside the [0,1] interval and
  decrease as regressors are added.}''
\end{quote}

While it is noteworthy that DA recognizes this, 
it is disappointing that it uses $R^{2}$ as it is known that this is not the best option for nonlinear models \cite{cameron1997, spiess2010}. Beyond this issue, the selection of alternative models appears largely random, underscoring the importance of user oversight in identifying appropriate models for the specific data and problem context. Moreover, our experiments suggest that the inherent structure of LLMs is not well-suited
for certain model-building tasks, such as hyperparameter tuning. In this respect, and in line with
the so-called Data Science quadrants \cite{Bie2022}, effective use of LLMs for automating
model-building tasks requires more targeted prompting, supplemented by technical expertise.

\subsubsection*{Back to Linear Regression}
In the following step, we asked ``What about simpler linear models with a smaller number of predictors?'' DA response provided some interesting comments on feature selection: 

\begin{mdframed}[backgroundcolor=grey!20, linewidth = 0pt] \vspace{0.1cm}
  \footnotesize 
\textbf{ChatGPT's DA} $[$...$]$\strut \hfill \Stars{5} \vspace{-0.4cm} \\ 
\begin{itemize}[itemsep=0pt, parsep=0pt, topsep=0pt, partopsep=0pt]
\item Univariate Analysis: Select features based on their individual correlation with the target variable.
\item Backward Elimination: Start with all variables and iteratively remove the least significant until the best subset is found.
\item Forward Selection: Start with no variables and iteratively add the most significant until the best subset is found.
\end{itemize}
\end{mdframed}

\noindent While the suggestion was sensible from a model selection viewpoint, the computations were interrupted due to an unexpected error in the platform. We kept the conversation flowing by asking ``Fit a simple regression model with one predictor only." 

\begin{mdframed}[backgroundcolor=grey!20, linewidth = 0pt] \vspace{0.1cm}
  \footnotesize 
\textbf{ChatGPT's DA} \strut \hfill \Stars{4} \vspace{-0.6cm}  \\ 
\begin{itemize}
\item Given these options, `area' often serves as a primary indicator of property price in real estate markets. We'll proceed with creating a simple linear regression model using `area' as the sole predictor for `price.'
\end{itemize}
\end{mdframed}

\begin{figure}[H]
\centering
\includegraphics[scale=0.33]{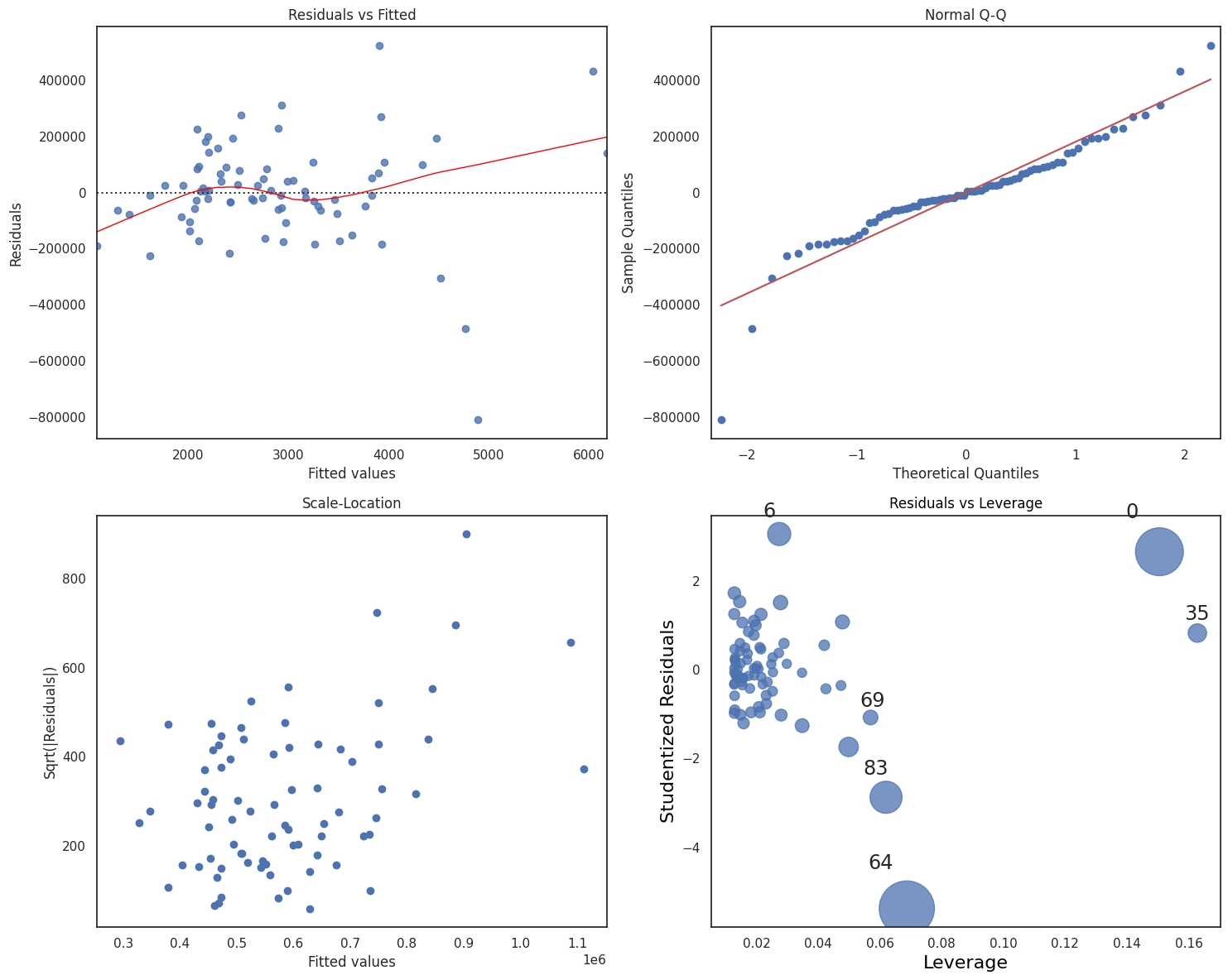}
\caption{\footnotesize Diagnostic plots for linear model for \texttt{duke\_forest} data (response: price; predictor: area).}
\footnotesize
\Stars{4}
\label{fig:simpleReg_ModelDiag}
\end{figure}

\noindent Apart from a few outputs on the fitted model (e.g., $R^2 = 0.398$), nothing was mentioned by DA about the  significance of the regression coefficients---nor even the values of the estimated coefficients themselves. After we asked ``Create a model diagnostic results for the simple regression with area,'' DA replied:

\begin{mdframed}[backgroundcolor=grey!20, linewidth = 0pt] \vspace{0.1cm}
  \footnotesize 
\textbf{ChatGPT's DA} $[$...$]$ \strut \hfill \Stars{5} \vspace{-0.4cm} \\ 
\begin{itemize}[itemsep=0pt, parsep=0pt, topsep=0pt, partopsep=0pt]
\item {Residuals vs. Fitted Plot: To check for non-linearity, unequal error variances, and outliers.}
\item {Normal Q-Q Plot: To assess if the residuals are normally distributed.}
\item {Scale-Location Plot: To verify the homoscedasticity of residuals (equal variance across the range of predictors).}
\item {Residuals vs. Leverage Plot: To identify influential data points.}
\end{itemize}
\end{mdframed}

\noindent and concluded with some sensible, yet shortsighted,   remarks:$^\dagger$\footnote{$^\dagger$For example, although the QQ plot of the residuals in Fig~\ref{fig:simpleReg_ModelDiag} may suggest a lack of normality, bands are not shown, nor a normality test is conducted---such as the well-known Jarque--Bera test \cite{jarque1987}.}

\begin{mdframed}[backgroundcolor=grey!20, linewidth = 0pt] \vspace{0.1cm}
  \footnotesize 
    \textbf{ChatGPT's DA} $[$...$]$ \strut \hfill \Stars{3}  \\ {These diagnostic plots suggest that while the model has some predictive power, there are violations of key assumptions like linearity, normality of residuals, and homoscedasticity. Such issues can affect the accuracy and reliability of the model's predictions. Exploring more complex models or transforming variables might help to address these issues.}
\end{mdframed}

The same prompt led to Fig~\ref{fig:simpleReg_ModelDiag} which is well conceived, apart from some minor aspects related with aesthetics—that could be easily edited in the raw Python code (e.g., labels outside of chart in the
studentized residuals). While we have not explicitly considered outliers in our prompts, it is well
known that the least squares estimator is highly sensitive to them. State-of-the-art methods from
robust statistics would be the most well-suited for that task (e.g., through standardized residuals
from the least trimmed squares estimator, \cite{rousseau2011}.

\subsection{Regression, II: Deep Neural Network}
In the next step, we asked ``Can you implement a neural network model based on your suggestion?'' DA wisely noted that:

\begin{mdframed}[backgroundcolor=grey!20, linewidth = 0pt] \vspace{0.1cm}
  \footnotesize 
  \textbf{ChatGPT's DA} \strut \hfill \Stars{5} \\ {Implementing a neural network model for real estate data is a more advanced approach that can potentially capture complex, non-linear relationships in the data.}
\end{mdframed}
and it suggested:

\begin{mdframed}[backgroundcolor=grey!20, linewidth = 0pt] \vspace{0.1cm}
  \footnotesize 
  \textbf{ChatGPT's DA} $[$...$]$\strut \hfill \Stars{5} \vspace{-0.4cm} \\ 
\begin{itemize}[itemsep=0pt, parsep=0pt, topsep=0pt, partopsep=0pt]
    \item Data Preprocessing:
\begin{itemize}[itemsep=0pt, parsep=0pt, topsep=0pt, partopsep=0pt]
    \item Scaling the features, since neural networks often perform better with normalized or standardized data.
    \item Using the same features (`bed', `bath', `area', `year\_built', `lot') and target variable (`price').
\end{itemize}
\item Building the Neural Network:
\begin{itemize}[itemsep=0pt, parsep=0pt, topsep=0pt, partopsep=0pt]
    \item We'll use a simple architecture with a few dense layers.
\end{itemize}
\item Training the Network:
\begin{itemize}[itemsep=0pt, parsep=0pt, topsep=0pt, partopsep=0pt]
    \item Fit the model to the training data.
\end{itemize}
\item Model Evaluation:
\begin{itemize}[itemsep=0pt, parsep=0pt, topsep=0pt, partopsep=0pt]
    \item Assess its performance on the test data.
\end{itemize}
\end{itemize}
\end{mdframed}

\noindent Clearly, the sample size ($n = 98$) might be too small for such a highly parameterized model, but to see how DA would perform we proceeded anyway. The suggested steps look informative, but the computation resulted in an error:

\begin{mdframed}[backgroundcolor=grey!20, linewidth = 0pt] \vspace{0.1cm}
  \footnotesize 
  \textbf{ChatGPT's DA} $[$...$]$ \strut \hfill \Stars{2} \\ {It appears that \textit{TensorFlow}, the library typically used for building neural network models, is not available in this environment. As a result, I'm unable to implement and train a neural network model directly here.}
\end{mdframed}
Yet, despite this error, DA found a solution and kept the conservation flowing by adding some code suggestions as a draft to use in a separate  environment (e.g., Jupyter Notebooks), given below:

\begin{mdframed}[backgroundcolor=grey!20, linewidth = 0pt] \vspace{0.1cm}
  \footnotesize 
  \textbf{ChatGPT's DA} $[$...$]$ \strut \hfill \Stars{3} \\ 
  {However, I can guide you through the process of building and training a simple neural network for this task using \textit{TensorFlow} and \textit{Keras} in your own environment. Here's a basic outline of how you could do it:} \vspace{-.4cm}
  \begin{center}\label{code1}
    [See Appendix A]
  \end{center}
\end{mdframed}

\noindent Much was left to the user’s discretion—for example, DA provided little guidance on the network
architecture, activation functions, loss function, or optimization algorithm for learning the weights.
On the positive side, the code is well documented, and the user can keep asking DA for further
details and interpretations.

\subsection{Regression, III: Bayesian Setting}
{In this section, we briefly explored Bayesian regression through direct clear prompting ("Fit a Bayesian Linear regression model using price as the response."); we specified no prior information, and used again the \texttt{Duke Forest} data set.} 

{DA returned: ``Bayesian model fitting took too long and was automatically interrupted due to computational complexity,'' along with simplification suggestions such as:}\vspace{0.2cm}

\begin{mdframed}[backgroundcolor=grey!20, linewidth = 0pt] \vspace{0.1cm}
  \footnotesize 
  \textbf{ChatGPT's DA} $[$...$]$ \strut \hfill \Stars{1} \\ 
  {To proceed efficiently, I’ll:
- Reduce the number of samples (for speed).
- Use a simpler prior or model specification to speed up compilation and sampling.} \vspace{-.4cm}
  \begin{center}\label{code1}
  \end{center}
\end{mdframed}

\vspace{0.2cm}
\noindent {However, DA did not succeed in this task, despite several attempts and requests from our side.} 

\section{Experiments on Unsupervised Data}\label{unsupervised}
In a separate session, we considered the implementation of unsupervised learning methods on the \texttt{duke\_forest} dataset. We started with the question ``Is the given data suitable for unsupervised learning?'' The generated response was informative for the first step:$^\dagger$\footnote{$^\dagger$The response by DA could inadvertently suggest that unsupervised learning is not feasible in the presence of missing values, which is a misconception as evidenced by \cite{ilin2010, chi2016, aschenbruck2023}.}

\begin{mdframed}[backgroundcolor=grey!20, linewidth = 0pt] \vspace{0.1cm}
  \footnotesize 
  \textbf{ChatGPT's DA} \strut \hfill \Stars{5}  \\ 
{To determine if the provided data is suitable for unsupervised learning, I will first analyze the contents of the file you've uploaded. Let's begin by examining the data in `duke\_forest.csv'. I'll check for:}
\begin{itemize}[itemsep=0pt, parsep=0pt, topsep=0pt, partopsep=0pt]
\item The nature of the data (numerical, categorical, etc.).
\item The presence of any missing values.
\item The overall structure and distribution of the data. 
\end{itemize}
{This analysis will help in understanding if the data is appropriate for unsupervised learning techniques such as clustering, dimensionality reduction, or anomaly detection. Let's take a look at the data now.}
\end{mdframed}

\noindent Also, potential use cases for unsupervised learning were nicely summarized as shown below:

\begin{mdframed}[backgroundcolor=grey!20, linewidth = 0pt] \vspace{0.1cm}
  \footnotesize 
\textbf{ChatGPT's DA} $[$...$]$\strut \hfill \Stars{5} \vspace{-0.4cm} \\ 
\begin{itemize}[itemsep=0pt, parsep=0pt, topsep=0pt, partopsep=0pt]
    \item {Clustering: To group properties into clusters based on similar characteristics.}
    \item {Dimensionality Reduction: To reduce the number of features while retaining important information, possibly for visualization.}
    \item {Anomaly Detection: To identify unusual properties, which could be outliers in terms of price or other features.}
\end{itemize}
{Given this information, the dataset seems suitable for unsupervised learning with a few considerations:}
\begin{itemize}[itemsep=0pt, parsep=0pt, topsep=0pt, partopsep=0pt]
    \item {Preprocessing: Handle missing values and encode categorical features.}
    \item {Appropriate Algorithms: Depending on the goal (e.g., clustering, dimensionality reduction), different algorithms can be applied. For instance, $k$-means clustering might be used to group similar properties, while PCA (Principal Component Analysis) could be useful for dimensionality reduction.}
    \item {Feature Selection: Given the mixed data types, careful selection of features and preprocessing (like normalization for numerical features) will be important.}
\end{itemize}
\end{mdframed}

\begin{figure}[H]
\centering
\includegraphics[scale=0.5]{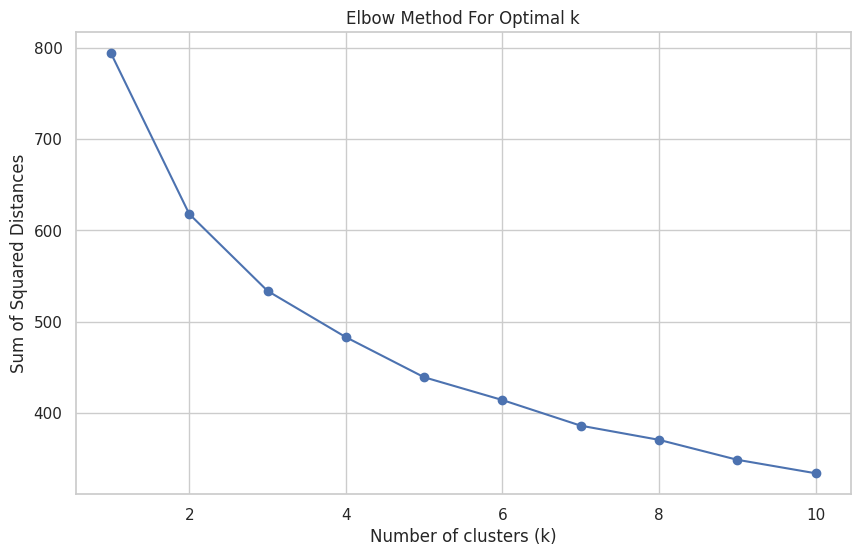}
\caption{\footnotesize Elbow method for \texttt{duke\_forest} data.}
\footnotesize
\Stars{5}
\label{fig:elbow_Kmeans}
\end{figure}

\noindent Among the different suggested algorithms we opted for clustering for the sake of illustration, and hence asked ``Proceed by implementing $k$-means
clustering.'' Next, DA automatically implements $k$-means clustering over a range of values and attempts to use the elbow method to determine an appropriate number of clusters ($k$). Supporting Information includes the Python code generated by DA to produce Fig~\ref{fig:elbow_Kmeans}.

Regarding the interpretation of the generated Elbow plot, shown in Figure~\ref{fig:elbow_Kmeans}, the interpretations were fairly reasonable:

\begin{mdframed}[backgroundcolor=grey!20, linewidth = 0pt] \vspace{0.1cm}
  \footnotesize
  \textbf{ChatGPT's DA} \strut \hfill \Stars{4}  \\ 
{Next, we'll use the Elbow Method to determine an appropriate number of clusters ($k$) for $k$-means clustering. This method involves running the $k$-means clustering for a range of $k$ values and calculating the Sum of Squared Distances (SSD) for each $k$. The ``elbow" point in the plot of SSD versus $k$ is considered as a good estimate for the number of clusters.}

{The Elbow Method plot shows the Sum of Squared Distances (SSD) for different numbers of clusters ($k$). In this plot, we are looking for a point where the SSD starts to decrease at a slower rate, forming an ``elbow." However, the plot does not show a very clear ``elbow," which can sometimes happen. In such cases, choosing the right number of clusters may require a bit of experimentation or may depend on additional domain knowledge or specific requirements}
\end{mdframed}

\section{Closing Remarks}\label{closing}
We have presented a hands-on review that examines the strengths and limitations of one of the
latest additions to the statistical software toolbox: OpenAI’s DA plugin. 

The history of statistical software is extensive and multifaceted, with various aspects detailed in
\cite{chambers2017}. Despite its current limitations, GPT has the potential to become a key player in
this landscape and to augment traditional statistical tools, providing researchers and practitioners
with unprecedented analytical capabilities.

Overall, the DA plugin can be useful for exploratory purposes. Yet, the modeling component
requires a detailed understanding and the application of more specific prompts along with domain
expertise. Without these, the generated outputs may be misleading. This is especially true for
novice learners, for whom the use of such a tool without meticulous supervision could inadvertently
hinder their learning process.

This note highlights the potential of the DA plugin to work as a co-pilot in Data Science workflows. While not the central focus here, the illustrations above demonstrate its potential to streamline and partially automate key steps in the process. At this stage, tools such as the DA plugin are
better suited to assistance-style automation rather than full automation. For certain Data Science
tasks, system-level prompting with targeted instructions may help address some of the issues noted
above. Yet, for more complex tasks, a multi-faceted and hybrid deployment of these tools is likely
required within an AI-augmented Data Science workflow.

In a well-known quote, world-renowned statistician Brian Ripley once claimed:
\begin{quote}\footnotesize 
\textsl{``Let's not kid ourselves: the most widely used piece of software for statistics is Excel.''} \\
  \strut \hfill \cite{ripley2002}.
\end{quote}

Since then, modern statistical computing tools—such as Julia, Python, and R—have been taking
over, but we might be reaching another paradigm shift. The interface between LLMs and statistical
models has been effectively established by ChatGPT’s DA plugin. Yet, as illustrated above, users
should critically assess its recommendations and outputs. Equipping the Data Science workflow with
appropriate computational components is essential to unlock new opportunities in automation. In
this regard, integrating multi-agent frameworks and leveraging established systems such as AutoML
may play a pioneering role in shaping the future of Data Science. Crucially, such efforts must involve
not only the use of LLMs but also sustained expert oversight

Our focus has been on illustrating the DA toolkit, but a full evaluation would require analyses
across datasets and repeated prompting. Recent studies emphasize the need for benchmark test
problems and performance metrics; some of this literature is summarized in Appendix A (\cite{He2023}, \cite{Hu2024}, \cite{Huang2023}, \cite{Zhang2025} \cite{Zhu2024}, but none
of these manuscripts focuses on the DA plugin. In contrast, we focused on zero-shot prompting
for statistical analysis and the effects of repeated prompting, leaving detailed benchmarking and
systematic performance assessment for future work. 

\clearpage
\subsubsection*{Disclosure Statement}
No potential conflict of interest was reported by the authors.

\subsubsection*{Data and Code Availability}
The two main data sets we focused on are publicly available. {All initial prompts and new data examples that we have tested after the reviewer suggestions are available upon request.}

\subsubsection*{ORCID}

\noindent Ozan Evkaya (https://orcid.org/0000-0002-5076-8144)
\noindent Miguel de Carvalho (https://orcid.org/0000-0003-3248-6984)

\subsubsection*{Funding}
This work was partially supported by the Generative AI Laboratory project “Large Language Models: Are They Ready for Data Science?”. Funding from CIDMA is also gratefully
acknowledged under projects UIDB/04106/2020 and UIDP/04106/2020
and https://doi.org/10.54499/UIDP/04106/2020.

\bibliographystyle{unsrtnat}
\bibliography{references}  


\clearpage
\appendix

\section*{Supporting Information}

\section*{{Appendix~A: Further Literature}}

{Some recent papers focus on how generative AI is being applied to the various Data Science tasks---including the impact on traditional workflows (from data wrangling and exploratory analysis to modeling and decision-making), as well as the challenges regarding accuracy, interpretability, and ethical use. Some relevant papers include:} \vspace{0.2cm}

{\textbf{(i) DataSciBench:} \citet{Zhang2025} provides a benchmark to evaluate LLM capabilities in Data Science, DataSciBench. Their framework comprises three key procedures: prompt definition and collection, response integration and validation, and LLM evaluation; they rely on six Data Science tasks, including: (i) Data Cleaning \& Preprocessing, (ii) Data Exploration \& Statistical Understanding, (iii)  Visualization, (iv) Predictive Modelling, (v) Statistical Modelling, (vi) Data Mining \& Pattern Recognition and (vii) Interpretability \& Report Generation.}\vspace{0.2cm}

{\textbf{(ii) Text2Analysis:} \citet{He2023}
proposed a Text2Analysis benchmark for tabular data analysis to conduct advanced analysis tasks that go beyond the SQL-compatible operations. They introduce five annotation methods to improve data quality, and worked on four Data Science tasks, including: (i) Rudimentary Operations (such as group by, aggregation, filter, and sort), (ii) Basic Insights incorporates value-based ranking, group attribution, time-series trends, monotonicity, outlier detection, and unimodality, (iii) Forecasting and (iv) Chart Generation.}\vspace{0.2cm}

{\textbf{(iii) Infiagent-dabench:} \citet{Hu2024}
introduced an InfiAgent-DABench benchmark for Data Science tasks, which requires agents to interact with an execution environment to end-to-end solve complex Data Science tasks. They focused on: (i) Comprehensive Data Preprocessing (12.0\%), (ii) Summary Statistics (24.0\%), (iii) Feature Engineering (13.3\%), (iv) Correlation Analysis (19.2\%), (v) Machine Learning (5.1\%), (vi) Distribution Analysis (17.1\%) and (vii) Outlier Detection (9.3\%).}\vspace{0.2cm}

{\textbf{(iv) DA-Code:} The DA-Code benchmark is designed by \citet{Huang2023} to assess LLMs on challenging, real, and diverse Data Science tasks. This study evaluated nine LLMs within an agent framework, considering tasks grouped under the following categories: (i) Data Wrangling, (ii) Machine Learning, (iii) Data Manipulation, (iv) Data Insights, (v) Visualization and (vi) Statistical Analysis.}\vspace{0.2cm}

{\textbf{(v) StatQA:} \citet{Zhu2024} introduced StatQA, a benchmark designed for statistical analysis tasks. StatQA comprises 11,623 examples tailored to evaluate LLMs' proficiency in statistical tasks and their applicability assessment capabilities. Overall, five typical categories of statistical tasks are created, ranging over (i) Correlation Analysis, (ii) Contingency Table Test, (iii) Distribution Compliance Test, (iv) Variance Test and (v) Descriptive Statistics.} 


\section*{Appendix~B: Selected Python Codes Provided by DA}

\subsection*{B.1.~Code for Fitting Neural Network Model}
\begin{lstlisting}[language=Python]
pip install tensorflow
import tensorflow as tf
from sklearn.preprocessing import StandardScaler
from sklearn.model_selection import train_test_split
from sklearn.metrics import mean_squared_error, r2_score

# Assuming X and y are your features and target variable

# Scale the features
scaler = StandardScaler()
X_scaled = scaler.fit_transform(X)
y_scaled = scaler.fit_transform(y.values.reshape(-1, 1))

# Split the data
X_train, X_test, y_train, y_test = train_test_split(X_scaled, y_scaled, 
test_size=0.2, 
random_state=0)

# Build the model
model = tf.keras.models.Sequential([
    tf.keras.layers.Dense(128, activation='relu', input_shape=(X_train.shape[1],)),
    tf.keras.layers.Dense(64, activation='relu'),
    tf.keras.layers.Dense(32, activation='relu'),
    tf.keras.layers.Dense(1)
])

# Compile the model
model.compile(optimizer='adam', loss='mean_squared_error')

# Train the model
model.fit(X_train, y_train, epochs=100, batch_size=10, verbose=0, validation_split=0.2)

# Evaluate the model
y_pred = model.predict(X_test)
mse = mean_squared_error(y_test, y_pred)
r2 = r2_score(y_test, y_pred)

print(f'MSE: {mse}, R-squared: {r2}')
\end{lstlisting}

\clearpage
\subsection*{A.2.~Code for Reproducing Fig.~\ref{fig:elbow_Kmeans}}

\begin{lstlisting}[language=Python]
from sklearn.preprocessing import StandardScaler, OneHotEncoder
from sklearn.compose import ColumnTransformer
from sklearn.impute import SimpleImputer
from sklearn.pipeline import Pipeline
import numpy as np

# Selecting relevant features for clustering
features_to_use = ['price', 'bed', 'bath', 'area', 'year_built', 'lot', 'type', 'heating', 
'cooling', 'parking']
data_clustering = data[features_to_use]

# Handling missing values
# Impute missing values in 'lot' with the median
# Drop `hoa' as it has too many missing values
data_clustering['lot'] = data_clustering['lot'].fillna(data_clustering['lot'].median())

# Categorical and numerical features
categorical_features = ['type', 'heating', 'cooling', 'parking']
numerical_features = ['price', 'bed', 'bath', 'area', 'year_built', 'lot']

# Creating a column transformer for preprocessing
preprocessor = ColumnTransformer(
    transformers=[
        ('num', StandardScaler(), numerical_features),
        ('cat', OneHotEncoder(handle_unknown='ignore'), categorical_features)
    ])

# Applying the transformations
data_preprocessed = preprocessor.fit_transform(data_clustering)

# Checking the shape of the preprocessed data
data_preprocessed.shape

from sklearn.cluster import KMeans
import matplotlib.pyplot as plt

# Range of k values to try
k_values = range(1, 11)

# List to store the sum of squared distances for each k
ssd = []

for k in k_values:
    kmeans = KMeans(n_clusters=k, random_state=0)
    kmeans.fit(data_preprocessed)
    ssd.append(kmeans.inertia_)

# Plotting the Elbow Method graph
plt.figure(figsize=(10, 6))
plt.plot(k_values, ssd, marker='o')
plt.title('Elbow Method For Optimal k')
plt.xlabel('Number of clusters (k)')
plt.ylabel('Sum of Squared Distances')
plt.grid(True)
plt.show()
\end{lstlisting}

\clearpage
\subsection*{Appendix~C: Prompting}
\label{prompting}
\subsection*{Impact of Prompting}  
The way prompts are phrased can strongly influence the analyses and recommendations. We illustrate this in a single prompt using again the \texttt{duke\_forest} dataset. Generally, from our experience, the given outputs are more structured when prompts are strategically oriented---such as ``Do necessary data cleaning if needed, and handle missing values:'' \vspace{0.2cm}

\begin{mdframed}[backgroundcolor=grey!20, linewidth = 0pt] \vspace{0.1cm}
  \footnotesize 
\textbf{ChatGPT's DA} \strut \hfill \Stars{3}  \\ 
{Here's a summary of the data cleaning and validation steps:}
\begin{itemize}[itemsep=0pt, parsep=0pt, topsep=0pt, partopsep=0pt]
\item {Data Structure and Missing Values:}
{The dataset contains 98 entries and 14 columns.
    There is one missing value in the `lot' column.
    The `hoa' column has 97 missing values out of 98 entries, indicating that it is largely empty and may not be useful for analysis.
    There are no duplicate rows in the dataset.}
\item {Removed Columns: The `Unnamed: 0' column, which was a redundant index, has been removed. The `hoa' column, which was largely empty, has also been removed.}
\item {Missing Value Treatment: The missing value in the `lot' column has been imputed with the median value of that column.}
\end{itemize}
\end{mdframed}\vspace{0.2cm}

\noindent Parenthetically, we underscore that the approach selected by DA to handle missing values (i.e., median imputation) is rather unsatisfactory from a statistical viewpoint. Methods such as stochastic regression imputation or multiple imputation offer more appropriate solutions to this issue (\cite{molenberghs2014, little2019}). \vspace{0.2cm}

{Finally, we note that \citet{He2023} investigates the role of clear versus unclear queries, showing that specific terminology in prompts is crucial because many user queries are ambiguous. Clear queries are crucial for obtaining reliable responses, underscoring the importance of user experience.} 

\subsection*{Selected Comments on Repeated Prompting}  
{Our experiments highlight the inherent randomness of LLMs such as ChatGPT's DA plug-in, showing that zero-shot prompts can yield variable results across tasks, even when repeated. While some tasks, like data visualization, produce outputs that are often unreliable or difficult to interpret, performance generally declines as task complexity increases, particularly when multiple steps are required. These findings emphasize the need for human oversight in automated data analysis, since certain parts of the workflow may require user confirmation or expert judgment, especially for interpretative tasks and visualizations, where uncontrolled automation risks misleading outcomes.}

\end{document}